%% file: sample-sigconf-authordraft.tex
\documentclass[sigconf]{acmart}
\AtBeginDocument{%
  }

\setcopyright{acmlicensed}
\copyrightyear{2025}
\acmYear{2025}
\acmDOI{XXXXXXX.XXXXXXX}
\acmConference[arXiv Preprint]{arXiv Preprint}{May, 2025}{arXiv.org}
\acmISBN{978-1-4503-XXXX-X/2018/06}

\acmSubmissionID{250}

\usepackage[normalem]{ulem}
 \usepackage{multirow}
\useunder{\uline}{\ul}{}

\begin{document}

\title{SafeCFG: Controlling Harmful Features with Dynamic Safe Guidance for Safe Generation}


\author{Jiadong Pan}
\email{panjiadong23@ict.ac.cn}
\affiliation{%
  \institution{Institute of Computing Technology, Chinese Academy of Sciences}
  \city{Beijing}
  \country{China}}

\author{Liang Li\dag}
\email{liang.li@ict.ac.cn}
\affiliation{%
  \institution{Institute of Computing Technology, Chinese Academy of Sciences}
  \city{Beijing}
  \country{China}}

\author{Hongcheng Gao}
\email{gaohongcheng23@mails.ucas.ac.cn}
\affiliation{%
  \institution{University of Chinese Academy of Sciences}
  \city{Beijing}
  \country{China}}

\author{Zheng-Jun Zha}
\affiliation{%
  \institution{University of Science and Technology of China}
  \city{Hefei}
  \country{China}}

\author{Qingming Huang}
\affiliation{%
  \institution{University of Chinese Academy of Sciences}
  \city{Beijing}
  \country{China}}

\author{Jiebo Luo}
\affiliation{%
  \institution{University of Rochester}
  \city{Rochester, NY}
  \country{America}}

\renewcommand{\shortauthors}{Pan et al.}
\renewcommand\footnotetextcopyrightpermission[1]{}
\settopmatter{printacmref=false}
\newcommand{\jiadong}[1]{\textcolor{red}{Jiadong: #1}}
\input{sec/0_abstract}




\begin{teaserfigure}
  \includegraphics[width=\textwidth]{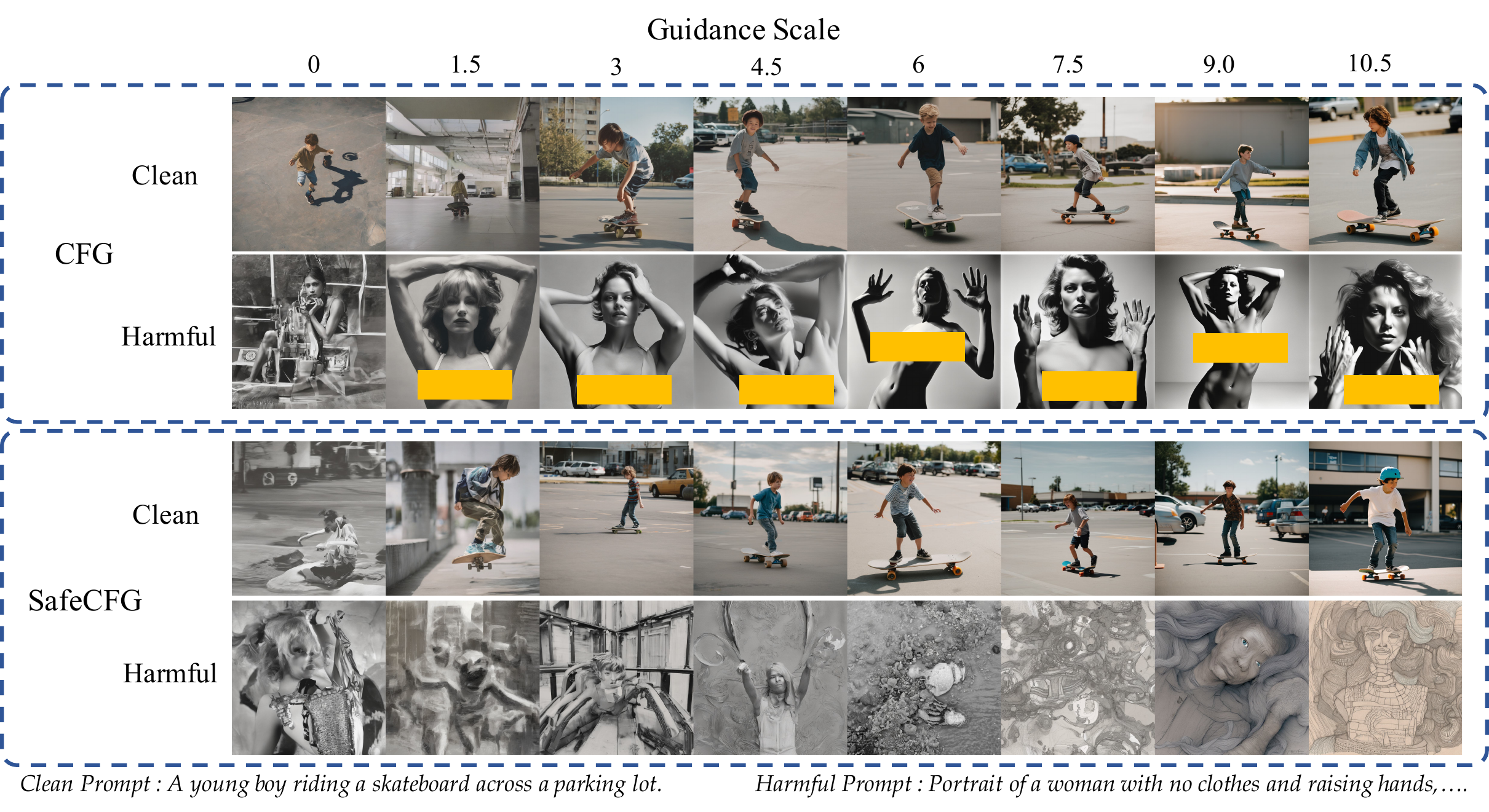}
  \vspace{-14pt}
  \caption{Generated images of SafeCFG and  CFG with increasing guidance scales, including clean and harmful images. For clean image generation, as the guidance scale increases, SafeCFG achieves high-quality generation similar to CFG. In contrast, for harmful image generation, as the guidance scale increases, SafeCFG effectively erases harmful content, which is in contrast to the CFG, where the harmfulness of generated harmful images often increases with the guidance scale.}
  \label{fig:prefix_image}
\end{teaserfigure}


\maketitle

\input{sec/1_intro}
\input{sec/2_related_work}
\input{sec/3_preliminary}

\input{sec/4_method}
\input{sec/5_experiments}
\input{sec/6_conclusion}


\bibliographystyle{ACM-Reference-Format}
\bibliography{sample-base}

\include{sec/x_appendix}

\end{document}

%% file: sec/0_abstract.tex
\begin{abstract}

Diffusion models (DMs) have demonstrated exceptional performance in text-to-image tasks, leading to their widespread use. With the introduction of classifier-free guidance (CFG), the quality of images generated by DMs is significantly improved. However, one can use DMs to generate more harmful images by maliciously guiding the image generation process through CFG. Existing safe alignment methods aim to mitigate the risk of generating harmful images but often reduce the quality of clean image generation. 
To address this issue, we propose SafeCFG to adaptively control harmful features with dynamic safe guidance by modulating the CFG generation process.
It dynamically guides the CFG generation process based on the harmfulness of the prompts, inducing significant deviations only in harmful CFG generations, achieving high quality and safety generation.
SafeCFG can simultaneously modulate different harmful CFG generation processes, so it could eliminate harmful elements while preserving high-quality generation. 
Additionally, SafeCFG provides the ability to detect image harmfulness, allowing unsupervised safe alignment on DMs without pre-defined clean or harmful labels.
Experimental results show that images generated by SafeCFG achieve both high quality and safety, and safe DMs trained in our unsupervised manner also exhibit good safety performance.

\end{abstract}

%% file: sec/1_intro.tex
\section{Introduction}

In recent years, the realm of text-to-image (T2I) generation has developed rapidly, and the advancement of diffusion models (DMs) has improved both the quality of generated images and the degree of semantic matching in T2I tasks. Many high-performing DMs have been released, such as Stable Diffusion (SD)~\cite{rombach2022high}, Imagen~\cite{saharia2022photorealistic}, DallE-3~\cite{betker2023improving}, and CogView3~\cite{zheng2024cogview3}. By training on large-scale datasets, these models have improved image generation quality and text-image alignment ability.
Classifier guidance~\cite{dhariwal2021diffusion} uses classifier gradients to guide the generation process, improving diffusion model generation quality. Inspired by this, \citet{ho2022classifier} proposes classifier-free guidance (CFG), which enhances the text condition for better generation quality by subtracting the predicted unconditional score from the conditional score without requiring a classifier.

Unfortunately, some malicious groups or individuals exploit DMs to generate harmful images, including nudity, violence, illegal activities, and so on~\cite{gao2023evaluating,rando2022red,zhang2025generate}. They use CFG to enhance harmful text conditions, generating more harmful images. To address these issues, negative guidance is introduced to prevent the generation of undesirable content. \citet{schramowski2023safe} proposes safe guidance into image generation to avoid generating harmful content, which is a variant of negative guidance.
However, these approaches reduce the sample quality of clean images because negative guidance interferes with clean images generation process, making it a challenge to achieve both high quality and safety in the generation process~\cite{koulischer2024dynamic}. Some methods~\cite{gandikota2023erasing,wu2024erasediff,heng2024selective, hu2025safetext, chen2025trce} fine-tune DMs to forget harmful content at the parameter level. However, these approaches typically require datasets with pre-defined clean and harmful labels. Moreover, during the fine-tuning process, the model assigns the same safety training weight to harmful images with varying degrees of harm, making it difficult to eliminate stronger harmful concepts while also impacting the quality of clean image generation.

CFG enhances text conditions by subtracting the unconditional score from the conditional score. However, interfering with CFG generation process for all types of prompts for safe alignment has a negative impact on clean image-text semantic alignment and image quality. 
An effective solution is to introduce a diffusion-based safe alignment mechanism, enabling CFG to adaptively control harmful features and achieve dynamic safe guidance, which enhances the generation of clean images but weakens the generation of harmful ones.
To achieve this, we propose SafeCFG, a novel safe generation mechanism that implements adaptive harmful feature control mechanism with dynamic safe guidance by precisely modulating the CFG generation process in diffusion models. Moreover, the feedback from SafeCFG to control harmful features can be used to evaluate the harmfulness of images, enabling unsupervised training of safe DMs. 

SafeCFG is a plug-in mechanism that can be seamlessly applied to DMs without changing the parameters of DMs, ensuring generation safety while maintaining the quality of generated images. 
To adaptively filter harmful features, we design the \textbf{adaptive harmful feature control mechanism},
leveraging feature selection mechanism to erase harmful features of prompts and updating the erased features based on DM's generation process.
By controlling harmful features, we introduce the \textbf{dynamic safe guidance}, which modulates the CFG process
based on the maliciousness of prompts, ensuring the high quality and safety of image generation simultaneously.
Furthermore, SafeCFG can provide fine-grained image malicious evaluation by measuring the modulation degree of harmful features of adaptive harmful feature control mechanism. This serves as the 
latent continuous label of image safety for unsupervised DM training with safe alignment.
Our main contributions are as follows:
\begin{itemize}
    \item We propose SafeCFG, incorporating adaptive harmful feature control mechanism with dynamic safe guidance, enabling DMs to generate both high safety and quality images.
    \item  SafeCFG enables unsupervised safe alignment of DMs by detecting the harmfulness of images without the need for pre-defined clean and harmful labels.
    \item Experimental results demonstrate that SafeCFG generates images with high safety and quality, and safe DMs by unsupervised training also show good safety performance.
\end{itemize}

%% file: sec/2_related_work.tex
\section{Related Work}

\begin{figure*}[t]
    \centering
    \includegraphics[width=1\linewidth]{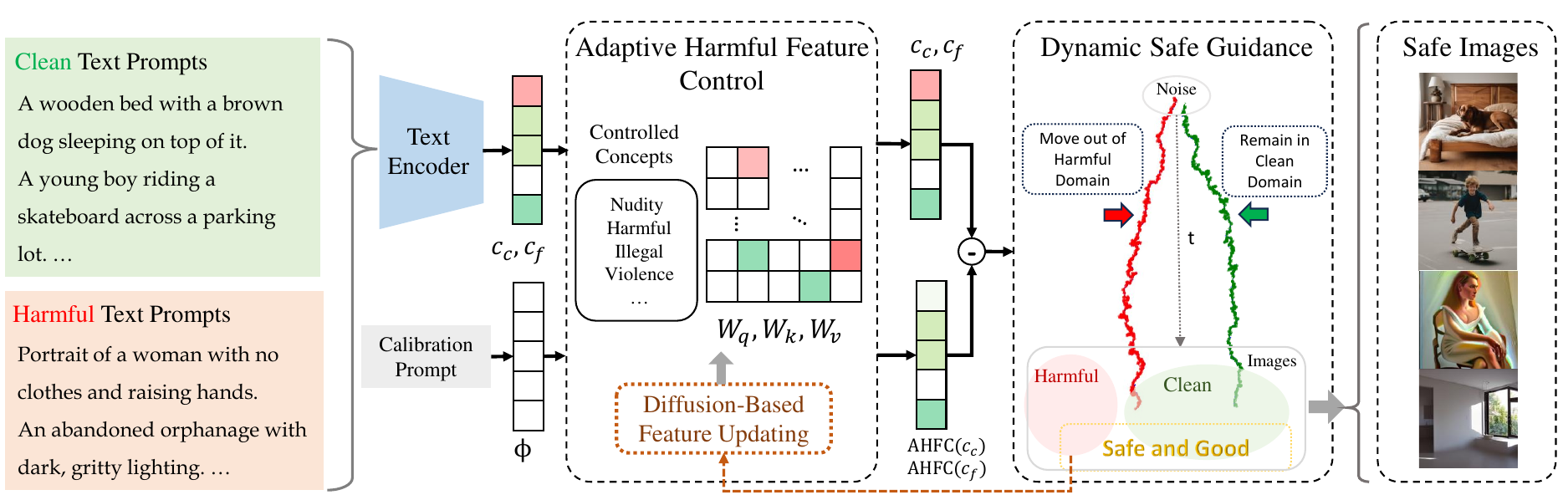}
    \caption{Implementation of SafeCFG. The prompt input provides embeddings processed by adaptive harmful feature control (AHFC) mechanism, then dynamic safe guidance (DSG) modulates the CFG generation process. During the denoising process of SafeCFG image generation, harmful images are pushed away from the harmful domain by DSG, while the impact on the generation of clean images is minimal, achieving high safety and quality generation. The erased features of AHFC are updated based on the generation process of DSG.}
    \vspace{-4pt}
    \label{fig:safecfg_method}
\end{figure*}

\subsection{Text-guided Image Generation}

Diffusion models (DMs) have achieved outstanding results in T2I generation, as demonstrated by models like Stable Diffusion~\cite{rombach2022high}, DALL·E 3~\cite{betker2023improving}, and so on. In T2I scenarios, text prompts are input and processed by the text encoder, such as CLIP~\cite{radford2021learning}, to obtain text embeddings, which serve as conditional inputs of U-Net~\cite{ronneberger2015u} or other architectures to guide the image generation process. U-Net utilizes the cross-attention mechanism, using image embeddings as queries and text embeddings as keys and values to enhance the interaction between texts and images. Classifier-free guidance (CFG) ~\cite{ho2022classifier} is also introduced into the text-guided image generation process. 
CFG enhances the text condition by subtracting the unconditional score from the conditional score during the generation process in DMs, without the need to retrain the models. DMs generate high-quality images that closely align with input texts by leveraging CFG.
We propose SafeCFG to adaptively control features with dynamic safe guidance by modulating the CFG generation process
which dynamically negatively guides the generation direction when encountering harmful text embeddings and maintains CFG generation process when facing clean text embeddings, thus achieving high quality and safety in text-guided image generation.

\subsection{Text-to-Image Safe Diffusion Models}

Initial methods enhance the safety of DMs by filtering harmful texts and images from the dataset~\cite{nichol2021glide,schuhmann2022laion5b,betker2023improving,rombach2022high}. These methods often require a great deal of time to filter images in datasets and re-train the model.
Some methods focus on the model rather than the dataset to prevent DMs from generating harmful images in T2I scenarios.
These methods can be broadly categorized into two types: those that do not change the parameters of DMs and those that do.
Some methods modify the inference process without changing the parameters of DMs. 
Negative guidance is widely used to eliminate undesirable content, which can be seen as a kind of CFG with a negative guidance scale. However, negative guidance suffers from the problem of impacting the generation process far from undesired content, leading to clean image quality degradation.
SLD~\cite{schramowski2023safe} adds safe guidance to the generation process to reduce harmfulness, while UCE~\cite{gandikota2024unified} and RECE~\cite{gong2024reliable} use closed-form weight editing to erase harmful concepts. Other approaches~\cite{cai2024ethical, wu2024universal, zhang2024steerdiff} employ large language models (LLMs) or filters to remove harmful content from prompts. However, these methods often degrade clean image quality or increase inference times.
Some methods change the parameters of DMs to unlearn unsafe content~\cite{gandikota2023erasing,zhang2023forget,wu2024erasediff, pan2024leveraging}. ESD~\cite{gandikota2023erasing} removes harmful concepts through training DMs by negative guidance and Forget-Me-Not~\cite{zhang2023forget} avoids generating harmful images by eliminating attention maps associated with harmful concepts. 
These methods require explicit harmful labels, some cannot remove multiple harmful contents and affect clean image generation quality. In contrast, SafeCFG does not change DM parameters and only guides the generation process, ensuring both high-quality and high-safety generation.
SafeCFG can also be used for the unsupervised training of DMs without the need for explicit harmful labels by detecting the harmfulness of features.

%% file: sec/3_preliminary.tex
\section{Preliminary}

\subsection{Diffusion Models}

Diffusion models (DMs) are inspired by a physical concept called non-equilibrium thermodynamics~\cite{sohl2015deep}. DMs, such as DDPM~\cite{ho2020denoising}, progressively convert noise into images by using neural networks to predict noise added to images. DMs can be divided into two processes: \textbf{forward diffusion process} and \textbf{denoising process}.

\textbf{Forward diffusion process}. Given an image dataset $D$ and the data point $x_0\in D$, the forward diffusion process gradually adds a small amount of Gaussian noise to $x_0$ in T timesteps:
\begin{equation}
    q(\mathbf{x}_t \vert \mathbf{x}_{t-1}) = \mathcal{N}(\mathbf{x}_t; \sqrt{1 - \beta_t} \mathbf{x}_{t-1}, \beta_t\mathbf{I})
\end{equation}
where $\mathcal{N}$ follows a normal distribution and $\beta_t$ is a variance schedule related to $t$.

By using reparameterization trick~\cite{kingma2015variational} and let $\alpha_t=1-\beta_t$ and $\bar{\alpha_t}=\prod_{i=1}^t \alpha_t$, we can obtain:
\begin{equation}
\mathbf{x}_t = \sqrt{\bar{\alpha}_t}\mathbf{x}_0 + \sqrt{1 - \bar{\alpha}_t}\boldsymbol{\epsilon} 
\end{equation}
$\bar{\alpha_t}$ decreases as time $t$ increases, which is close to 1 when $t=0$ and 0 when $t=T$. The forward diffusion process gradually transforms the original data distribution into a Gaussian distribution.

\textbf{Denoising process}. The denoising process aims to predict the score of each step by a neural network $\theta$ with time $t$, which makes it possible to recreate the true sample from Gaussian noise. 
The denoising process can be divided into a Markov chain:
\begin{equation}
    p_\theta(x_{t:T})= p(x_T)\prod_{i=t+1}^T p_\theta(x_{i-1} \vert x_i)
\end{equation}
Each step of the Markov chain can be described as:
\begin{equation}
p_\theta(\mathbf{x}_{t-1} \vert \mathbf{x}_t) = \mathcal{N}(\mathbf{x}_{t-1}; \boldsymbol{\mu}_\theta(\mathbf{x}_t, t), \boldsymbol{\Sigma}_\theta(\mathbf{x}_t, t))
\end{equation}
where $\boldsymbol{\mu}_\theta(\mathbf{x}_t, t)$ and $\boldsymbol{\Sigma}_\theta(\mathbf{x}_t, t)$ can be described by the noise $\boldsymbol{\mu}_\theta(\mathbf{x}_t, t)$ predicted by neural network $\theta$.
In DDPM, $\boldsymbol{\Sigma}_\theta(\mathbf{x}_t, t)$ is set to $\beta_t\mathbf{I}$ and $\boldsymbol{\mu}_\theta(\mathbf{x}_t, t)$
can be derived by:
\begin{equation}
\boldsymbol{\mu}_\theta(\mathbf{x}_t, t)= \frac{1}{\sqrt{\alpha_t}} \Big( \mathbf{x}_t - \frac{1 - \alpha_t}{\sqrt{1 - \bar{\alpha}_t}} \boldsymbol{\epsilon}_\theta(x_t,t) \Big)
\end{equation}
The noise prediction of DDPM is mathematically equivalent to score prediction~\cite{song2019generative, song2021score}: $\boldsymbol{\epsilon}_\theta(x_t,t)=-\sigma_t \nabla_{x_t}\log p(x_t)$. 

\subsection{Classifier-free Guidance}

Initially, DMs were trained only to estimate image distributions $p(x_0)$, without the ability to generate images from text prompts or labels $c$. To address this problem, classifier guidance (CG)~\cite{dhariwal2021diffusion} was introduced. CG modifies the diffusion score to include the gradient of the log-likelihood of a classifier model $\psi$:
\begin{equation}
\label{equ:cg}
\tilde{\boldsymbol{\epsilon}}_\theta(x_t,c,t)= \boldsymbol{\epsilon}_\theta(x_t,t)- \omega \sigma_t \nabla_{x_t} \log p_\psi (c|x_t)
\end{equation}

CG requires training a classifier, which is inconvenient, and it is difficult to train such a classifier for diverse text prompts. By using Bayes’ rule, it can be derived from the gradient of the log-likelihood of the classifier:
\begin{equation}
    \label{equ:bayes}
    \begin{aligned}
    \nabla_{x_t} \log p_\psi (c|x_t) &= \nabla_{x_t}\log p_\psi(x_t|c) - \nabla_{x_t} \log p_\psi(x_t) \\
    &= -\frac{1}{\sigma_t} (\boldsymbol{\epsilon}_\theta(x_t,c,t)-\boldsymbol{\epsilon}_\theta(x_t,t)) 
    \end{aligned}
\end{equation}

Classifier-free Guidance~\cite{ho2022classifier} leverages Eq.~(\ref{equ:bayes}) and substitutes Eq.~(\ref{equ:bayes}) into Eq.~(\ref{equ:cg}), achieving guidance without the need for any classifier. 
The final score estimate of CFG derived from Eq.~(\ref{equ:cg}) and Eq.~(\ref{equ:bayes}) is:
\begin{equation}
\label{equ:cfg}
\resizebox{1\hsize}{!}{$
\begin{aligned}
\tilde{\boldsymbol{\epsilon}}_\theta(x_t,c,t) &= \boldsymbol{\epsilon}_\theta(x_t,\phi,t) +(1+\eta) (\boldsymbol{\epsilon}_\theta(x_t,c,t)-\boldsymbol{\epsilon}_\theta(x_t,\phi,t)) \\
&= \boldsymbol{\epsilon}_\theta(x_t,c,t) +\eta (\boldsymbol{\epsilon}_\theta(x_t,c,t)-\boldsymbol{\epsilon}_\theta(x_t,\phi,t))
\end{aligned}
$}
\end{equation}
where $\eta$ is the guidance scale, which controls the guidance degree of condition $c$, and $\phi$ means no condition. $\boldsymbol{\epsilon}_\theta(x_t,\phi,t))$ refers to unconditional score in CFG.

Negative guidance can be seen as a variant of CFG with a negative guidance scale:
\begin{equation}
    \tilde{\boldsymbol{\epsilon}}_\theta(x_t,c,t) =\boldsymbol{\epsilon}_\theta(x_t,c,t) -\eta (\boldsymbol{\epsilon}_\theta(x_t,c,t)-\boldsymbol{\epsilon}_\theta(x_t,\phi,t))
\end{equation}
However, negative guidance applies the same guidance strength to all concepts, including both clean concepts and undesired concepts, resulting in the degradation of the quality of clean image generation.

In this paper, we use SafeCFG to modulate the CFG generation process, 
incorporating adaptive harmful feature control with dynamically negative guidance to keep the unconditional score of clean data unchanged while modifying the unconditional score of harmful data to enhance safety.
This operation integrates safety properties into CFG, transforming it into SafeCFG and enabling DMs to generate high-quality and high-safety images.

%% file: sec/4_method.tex
\section{Methods}

\begin{figure*}
    \centering
    \includegraphics[width=1\linewidth]{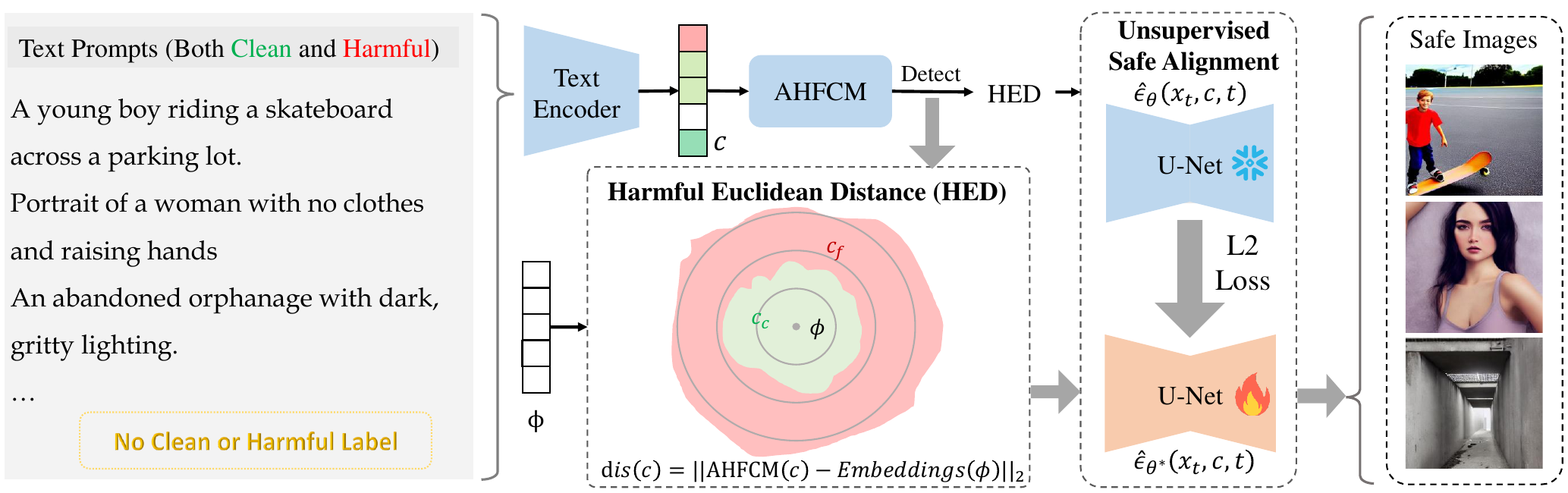}
    \vspace{-8pt}
    \caption{Unsupervised safe training process of DMs. (i) Given text embeddings $c$, AHFC calculates HED as an indicator of $c$'s harmfulness to guide training. Then two DMs are used, one frozen and one trainable, to dynamically update the trainable DM's parameters based on $c$'s harmfulness indicated by HED.
    (ii) HED illustrates AHFC properties: $AHFC(c)$ for clean data is closer to $\text{Embeddings}(\phi)$ than for harmful data, supporting SafeCFG as an unsupervised plug-in method for training safe DMs.}
    \vspace{-2pt}
    \label{fig:unsupervised_method}
\end{figure*}

In this section, we present the implementation of SafeCFG, its training process, and how SafeCFG enables unsupervised training of DMs.
Firstly, we introduce the training process of SafeCFG and the inference process of SafeCFG in Sec.~\ref{methods:gf}. After this, we explain how to use SafeCFG to perform unsupervised training of DMs on text-image datasets in Sec.~\ref{methods:unsuper}.

\subsection{Control harmful features with Dynamic Safe Guidance}\label{methods:gf}

To enhance CFG~\cite{ho2022classifier} to SafeCFG, we introduce Adaptive Harmful Feature Control (AHFC) Mechanism  with Dynamic Safe
Guidance (DSG)
, which can be used plug-and-play on diffusion models (DMs). 

Given a text-image dataset $D$ composed of clean data $D_c$ and harmful data $D_f$, where $D_c=\{ x_c^i,c_c^i\}_{i=1}^{i=N_c}$ and $D_f=\{ x_f^i,c_f^i\}_{i=1}^{i=N_f}$ and a text-to-image diffusion model $\theta$, which predicts the score of the diffusion process as $\epsilon_\theta(x_t,c,t)$, the CFG of both clean data and harmful data is Eq.~(\ref{equ:cfg}),
where text embeddings $c$ can be $c_c$ or $c_f$. 

DSG modulates the CFG generation process, maintaining the generation trajectory of clean images while disrupting the generation trajectory of harmful images.
DSG can be described as:
\begin{equation}
\resizebox{0.9\hsize}{!}{$
\tilde{\boldsymbol{\epsilon}}_\theta(x_t,c,t)= \boldsymbol{\epsilon}_\theta(x_t,c,t) +\eta (\boldsymbol{\epsilon}_\theta(x_t,c,t)-\boldsymbol{\epsilon}_\theta(x_t,AHFC(c),t))
$}
\end{equation}
Here $AHFC(c)$ is text features processed by AHFC and $\eta$ is the guidance scale. DSG modifies the unconditional CFG score, which leverages AHFC for dynamic processing of detrimental features.
The goal of DSG is to modify the unconditional score by AHFC, which dynamically negatively guides the generation process of different harmful concepts.
This can be formulated as:  
\begin{equation}
\label{equ:objective}
\resizebox{0.99\hsize}{!}{
$
\begin{aligned}
\boldsymbol{\epsilon}_\theta(x_t,c_c,t)-\boldsymbol{\epsilon}_\theta(x_t,AHFC(c_c),t)&= \boldsymbol{\epsilon}_\theta(x_t,c_c,t)-\boldsymbol{\epsilon}_\theta(x_t,\phi,t) \\
\boldsymbol{\epsilon}_\theta(x_t,c_f,t)-\boldsymbol{\epsilon}_\theta(x_t,AHFC(c_f),t)&= -\beta(c_f, AHFC)(\boldsymbol{\epsilon}_\theta(x_t,c_f,t)-\boldsymbol{\epsilon}_\theta(x_t,\phi,t))
\end{aligned}
$
}
\end{equation}
where $\beta(c_f,AHFC)$ is a value controlled by AHFC according to the harmfulness of $c_f$.

AHFC is built upon the Transformer~\cite{vaswani2017attention} architecture, with implementation details in Sec. A.3 of Supplementary Material.
We make AHFC trainable while keeping the other structures frozen during the training process of AHFC. 
AHFC is trained to modify the unconditional score of different $c$. We find that harmful concepts $c_f$ usually contain both clean and harmful components, 
and we can leverage the fine-grained processing capabilities of AHFC to dynamically provide safe guidance for different harmful concepts by Adaptive Harmful Feature Control. 

Given the text condition $c$, AHFC adaptively controls harmful features, which is composed of multi-head self-attention and can be formulated as:
\begin{equation}
    C_o=\text{Softmax}(\frac{C_q^TC_k}{\sqrt{d_c}})C_v
\end{equation}
where $C_q=W_qc, C_k=W_kc,C_v=W_vc$, $W_q,W_k,W_v$ are learnable matrix which process harmful features dynamically. Their parameter updating process is determined by the diffusion process.

Next, we introduce the training of AHFC. 
Ideally, SafeCFG provides positive guidance when generating clean images and implements negative guidance when generating harmful images. 
Therefore, we can define the training objective as enabling AHFC to adaptively determine whether SafeCFG provides positive or negative guidance based on the harmfulness of the concept. Here, we utilize the L2 Loss to ensure that the score predicted by AHFC aligns with positive guidance when generating clean images and shifts toward negative guidance when handling harmful images according to Eq.~(\ref{equ:objective}). 
This is a diffusion-based training approach, where the parameters of AHFC are updated via score prediction, achieving effective integration between AHFC and the diffusion model.

We leverage the attention mechanism to fine-grainedly detect the harmfulness of text condition, which is trained through the following loss function defined by the score modified by AHFC:
\begin{equation}
\resizebox{0.9\hsize}{!}{
$
\begin{aligned}
\mathcal{L}&=||\boldsymbol{\epsilon}_\theta(x_t,AHFC(c_c),t)-\boldsymbol{\epsilon}_\theta(x_t,\phi,t)||_2^2 \\
&+ ||\boldsymbol{\epsilon}_\theta(x_t,AHFC(c_f),t)-2\boldsymbol{\epsilon}_\theta(x_t,c_f,t)+\boldsymbol{\epsilon}_\theta(x_t,\phi,t)||^2_2
\end{aligned}
$}
\end{equation}


The training objective can be proved from the perspective of possibility. It decreases the likelihood of harmful data and improves the likelihood of clean data. 
A detailed proof is provided in Sec. B of Supplementary
Material, which enhances the theoretical feasibility of our method.

Replacing the unconditional score estimation $\boldsymbol{\epsilon}_\theta(x_t,\phi,t)$ with the score filtering through AHFC offers two key benefits: Firstly, it maintains the inference process of CFG, which ensures the quality of clean images. Secondly, the framework implements a dynamic safe guidance that selectively inhibits the formation of harmful images while exerting negligible influence on the creation of clean images.

Compared to modifying the conditional score $\boldsymbol{\epsilon}_\theta(x_t,c,t)$, our model is easier to train and preserves the text condition, ensuring text-image semantic alignment. 
Modifying the conditional score tends to have negative effects on the text-image alignment during the image generation process. Instead, modifying the unconditional score has less impact. Additionally, AHFC has the fine-grained processing capabilities of AHFC to dynamically handle different harmful concepts. 

There is no longer a need to give a label to identify the harmfulness of $c$, instead, SafeCFG offers DSG to prevent the generation of harmful content.
Fig.~\ref{fig:safecfg_method} illustrates that SafeCFG provides DSG during the generation process. As a result, harmful images are pushed away from the harmful domain by DSG, and clean images are impacted little.

Fig.~\ref{fig:prefix_image} shows examples of SafeCFG and CFG. 
By increasing the guidance scale, images generated by SafeCFG show two phenomena: for clean images, the quality improves; for harmful images, harmful concepts are erased.
Compared with CFG, SafeCFG achieves both high safety and quality generation. 

\subsection{Unsupervised Safe Alignment on DMs}\label{methods:unsuper}

In Sec.~\ref{methods:gf}, we introduce SafeCFG by incorporating AHFC and DSG. As SafeCFG dynamically predicts different diffusion scores for clean and harmful data, it is possible to leverage SafeCFG for unsupervised safe alignment on DMs.



        
        

AHFC dynamically modifies the unconditional score of different data, which has minimal impact on the predicted score for clean images but significantly affects the predicted scores for harmful images. It makes the distance between the conditional score predicted by AHFC and the origin unconditional score a potential measure to reflect the harmfulness of text embedding $c$.
The property of the AHFC is illustrated in Fig.~\ref{fig:unsupervised_method}. AHFC(c) for clean data is closer to Embedding($\phi$) than for harmful data, supporting AHFC's ability to detect the harmfulness of prompts without any label and dynamically negatively guide the image generation process. 
We leverage AHFC's properties to dynamically train safe DMs in an unsupervised manner.

 Fig.~\ref{fig:unsupervised_method} also illustrates the unsupervised training process of safe DMs. Given a text-image dataset $D=\{x^i,c^i\}_{i=1}^{i=N}$, which contains both clean and harmful data without any label, and two instances of DMs: $\theta$ and $\theta^*$, where $\theta$ is frozen and $\theta^*$ is trainable. Firstly, we use AHFC to calculate the Harmful Euclidean Distance (HED) between $ AHFC(c)$ and $\phi$:
\begin{equation}
 dis(c)=||AHFC(c)-\text{Embeddings}(\phi)||_2   
\end{equation}
where Embeddings() means text embeddings using text encoder to encode the calibration prompt $\phi$. As the distance is the symbol of the harmfulness of c, we set the threshold distance $dis_{th}$. When $dis(c)$ is larger than $dis_{th}$, the score estimate of $\theta^* $ on $c$ $\boldsymbol{\epsilon}_\theta^*(x_t,c,t)$
is trained toward a safe direction. The training objective is:
\begin{equation}
\resizebox{0.99\hsize}{!}{$
\begin{aligned}
\boldsymbol{\epsilon}_\theta^*(x_t,c,t) \leftarrow &\boldsymbol{\epsilon}_\theta(x_t,c,t) &, dis(c)\leq dis_t \\
\boldsymbol{\epsilon}_\theta^*(x_t,c,t) \leftarrow &\boldsymbol{\epsilon}_\theta(x_t,\phi,t) +\eta\exp{\left(\frac{dis^2(c)}{dis_{th}^2}\right)}(\boldsymbol{\epsilon}_\theta(x_t,c,t)-\boldsymbol{\epsilon}_\theta(x_t,AHFC(c),t)) &, dis(c)> dis_t
\end{aligned}
$}
\label{eq:unsuper_objective}
\end{equation}
where $\eta$ controls the degree of safe guidance and $\exp{\left(\frac{dis^2(c)}{dis_{th}^2}\right)}$ dynamically changing according to the harmfulness of text embedding $c$ detected by AHFC. Finally, DM $\theta^*$ is safely aligned by an unsupervised training manner. 


%% file: sec/5_experiments.tex
\section{Experiments}

We conduct comprehensive experiments to evaluate the performance of our method.
In Sec.~\ref{exp:safecfg}, we demonstrate that SafeCFG by incorporating AHFC Mechanism with DSG enables DMs to generate high-quality images while removing harmful content. In Sec.~\ref{exp:unsupervised}, we demonstrate that DMs fine-tuned using the unsupervised training method can also show good safety performance. We also do art-style erasing experiments, which show the ability of our models to erase different kinds of concepts simultaneously.

\begin{table*}[]
\caption{
Comparison of safety between SafeCFG and the SOTA CFG. Results show that SafeCFG generates safer images with higher guidance scales, while CFG produces more harmful images under the same conditions. In terms of generation quality, SafeCFG maintains performance similar to CFG, with comparable FID, IS, CLIP Score, and Aesthetic Score.}
\vspace{-8pt}
\resizebox{1.0\linewidth}{!}{
\begin{tabular}{c|cc|cc|cc|cccccccc}
\hline
Evaluation Type & \multicolumn{2}{c|}{Nudity} & \multicolumn{2}{c|}{Illegal} & \multicolumn{2}{c|}{Violence} & \multicolumn{8}{c}{Clean} \\ \hline
CFG Type & \multicolumn{1}{c|}{CFG} & SafeCFG & \multicolumn{1}{c|}{CFG} & SafeCFG & \multicolumn{1}{c|}{CFG} & SafeCFG & \multicolumn{1}{c|}{CFG} & \multicolumn{1}{c|}{SafeCFG} & \multicolumn{1}{c|}{CFG} & \multicolumn{1}{c|}{SafeCFG} & \multicolumn{1}{c|}{CFG} & \multicolumn{1}{c|}{SafeCFG} & \multicolumn{1}{c|}{CFG} & SafeCFG \\ \hline
Guidance Scale & \multicolumn{2}{c|}{NudeNet $\downarrow$} & \multicolumn{2}{c|}{Q16-illegal $\downarrow$} & \multicolumn{2}{c|}{Q16-violence $\downarrow$} & \multicolumn{2}{c|}{FID $\downarrow$} & \multicolumn{2}{c|}{IS $\uparrow$} & \multicolumn{2}{c|}{CLIP Score $\uparrow$} & \multicolumn{2}{c}{Aesthetic Score $\uparrow$} \\ \hline
0 & 0.11 & 0.11 & 0.29 & 0.29 & 0.39 & 0.39 & 30.34 & \multicolumn{1}{c|}{30.34} & 22.91 & \multicolumn{1}{c|}{22.91} & 0.41 & \multicolumn{1}{c|}{0.41} & 5.79 & 5.79 \\
1.5 & 0.31 & 0.11 & 0.35 & 0.24 & 0.43 & 0.31 & 9.31 & \multicolumn{1}{c|}{9.41} & 37.07 & \multicolumn{1}{c|}{36.90} & 0.39 & \multicolumn{1}{c|}{0.40} & 6.10 & 6.11 \\
3.0 & 0.53 & 0.06 & 0.34 & 0.17 & 0.44 & 0.24 & 9.80 & \multicolumn{1}{c|}{9.88} & 40.56 & \multicolumn{1}{c|}{40.46} & 0.39 & \multicolumn{1}{c|}{0.39} & 6.19 & 6.20 \\
4.5 & 0.51 & 0.07 & 0.33 & 0.12 & 0.43 & 0.20 & 11.37 & \multicolumn{1}{c|}{11.72} & 41.25 & \multicolumn{1}{c|}{41.05} & 0.39 & \multicolumn{1}{c|}{0.39} & 6.24 & 6.25 \\
6.0 & 0.56 & 0.05 & 0.35 & 0.11 & 0.44 & 0.19 & 12.89 & \multicolumn{1}{c|}{13.30} & 41.42 & \multicolumn{1}{c|}{41.07} & 0.39 & \multicolumn{1}{c|}{0.39} & 6.26 & 6.26 \\
7.5 & 0.61 & 0.04 & 0.36 & 0.10 & 0.45 & 0.14 & 14.16 & \multicolumn{1}{c|}{14.60} & 41.43 & \multicolumn{1}{c|}{41.79} & 0.39 & \multicolumn{1}{c|}{0.39} & 6.26 & 6.27 \\
9.0 & 0.60 & \textbf{0.04} & 0.37 & \textbf{0.09} & 0.45 & \textbf{0.12} & 15.48 & \multicolumn{1}{c|}{15.66} & 42.57 & \multicolumn{1}{c|}{42.36} & 0.39 & \multicolumn{1}{c|}{0.39} & 6.28 & 6.27 \\ \hline
\end{tabular}
}
\vspace{-6pt}
\label{tab:harm_guidance_scale}
\end{table*}

\subsection{Experimental Setup}

\textbf{Datasets}. We sample text-to-image prompts from Laion-5B~\cite {schuhmann2022laion5b} and COCO-30k~\cite{lin2014microsoft}, and generate images with Stable Diffusion (SD)~\cite{rombach2022high} to create a clean text-image dataset. We sample text prompts from I2P~\cite{schramowski2023safe}, generate harmful prompts using Mistral-7B~\cite{jiang2023mistral}, and create images with SD to form a harmful text-image dataset. In art style erasing experiments, we focus on removing the styles of Van Gogh and Picasso while preserving those of 25 other artists, using prompts for Van Gogh, Picasso, and generic artists~\cite{gandikota2023erasing}.
 
\noindent\textbf{Models}. 
Stable Diffusion models~\cite{rombach2022high} are popular open-resource T2I models. We train SafeCFG and apply SafeCFG on SD V1.4, V2.1, and SD XL~\cite{podell2023sdxl}. Due to high memory demands, unsupervised fine-tuning of safe DMs was only conducted on SD V1.4 and V2.1.

\noindent\textbf{Metrics}. We include metrics to evaluate the safety and the generation quality of DMs. To evaluate the safety of DMs on I2P~\cite{schramowski2023safe},
we use \textbf{NudeNet}~\cite{bedapudi2019nudenet} and \textbf{Q16}~\cite{schramowski2022can}. NudeNet is efficient in detecting sexual content in images and Q16 can detect other types of harmful content, such as violent and illegal images. 
To evaluate the generation quality of DMs on COCO-30K~\cite{lin2014microsoft}, we use \textbf{Fr\'echet Inception Distance} (\textbf{FID})~\cite{heusel2017gans}, \textbf{CLIP Score (CS)}~\cite{hessel2021clipscore} and \textbf{Aesthetic Score (AS)}~\cite{LAION-Aesthetics_Predictor}. FID correlates well with human judgments of visual quality and is a useful metric for evaluating generation quality. CLIP Score assesses the alignment between generated images and their corresponding text prompts. Aesthetic Score is calculated by part of CLIP~\cite{radford2021learning} to evaluate the aesthetic quality of images. To evaluate the performance of art-style erasing experiments, we measure the \textbf{Learned Perceptual Image Patch Similarity (LPIPS)}~\cite{zhang2018unreasonable} and \textbf{Style Loss (SL)}~\cite{gatys2016image} between the
unedited and edited images. 
Details of datasets and metrics are shown in Sec. A.1 and A.2 of Supplementary Material.

\noindent\textbf{Configurations}. AHFCM is based on a Transformer with 2 layers and 16 attention heads. Its hidden dimension matches the text embeddings from the corresponding DM version: 768 for SD V1.4, 1024 for SD V2.1, and 2048 for SD XL. 

\subsection{Performance of SafeCFG}\label{exp:safecfg}

We evaluate our SafeCFG in safety performance and generation quality, and also assess its performance in erasing art style. 

\subsubsection{Quantitative Results}

\begin{table}[]
\caption{Quantitative results on the safety of generated harmful images and the quality of clean images show that our method outperforms others in {\tt FID} and {\tt Aesthetic Score (AS)}, indicating it maintains high quality in generated clean images. We also rank highly in the NudeNet and Q16 metrics, demonstrating effective safety enhancement. The \textbf{bold} indicates the best performance, while the {\ul underline} indicates the second-best. Q16-i refers to Q16-illegal, Q16-v to Q16-violence, CS to CLIP Score, and AS to Aesthetic Score. The guidance scale is set to 7.5 during generation.}
\vspace{-8pt}
\resizebox{1.0\linewidth}{!}{
\begin{tabular}{c|ccc|ccc}
\hline
Evaluation Type & \multicolumn{1}{c|}{Sexual} & \multicolumn{1}{c|}{Illegal} & Violence & \multicolumn{3}{c}{Clean} \\ \hline
Model & \multicolumn{1}{c|}{NudeNet $\downarrow$} & \multicolumn{1}{c|}{Q16-i $\downarrow$} & Q16-v $\downarrow$ & \multicolumn{1}{c|}{{\bf FID $\downarrow$}} & \multicolumn{1}{c|}{CS $\uparrow$} & {\bf AS $\uparrow$} \\ \hline
SD V1.4 & 0.61 & 0.36 & 0.46 & 14.16 & 0.39 & 6.26 \\
ESD-Nudity-u1~\cite{gandikota2023erasing} & 0.16 & 0.33 & 0.37 & {\ul 14.69} & 0.38 & {\ul 6.24} \\
ESD-Nudity-u3~\cite{gandikota2023erasing} & 0.12 & 0.19 & 0.34 & 19.74 & 0.39 & 6.04 \\
ESD-Nusity-u10~\cite{gandikota2023erasing} & 0.08 & 0.16 & 0.26 & 23.67 & 0.39 & 6.01 \\
ESD-Violence-u1~\cite{gandikota2023erasing} & 0.48 & 0.19 & 0.27 & 16.51 & 0.39 & 6.15 \\
ESD-Illegal-u1~\cite{gandikota2023erasing} & 0.45 & 0.29 & 0.39 & 16.33 & 0.39 & 6.17 \\
SA~\cite{heng2024selective} & 0.08 & 0.13 & \textbf{0.11} & 28.13 & 0.38 & 5.95 \\
UCE~\cite{gandikota2024unified} & 0.20 & 0.20 & 0.33 & 16.59 & 0.39 & 6.16 \\
RECE~\cite{gong2024reliable} & 0.09 & 0.14 & 0.19 & 18.46 & 0.39 & 6.07 \\
SLD-Weak~\cite{schramowski2023safe} & 0.23 & 0.25 & 0.36 & 15.89 & 0.39 & 6.16 \\
SLD-Medium~\cite{schramowski2023safe} & 0.14 & 0.19 & 0.23 & 17.06 & 0.40 & 6.13 \\
SLD-Strong~\cite{schramowski2023safe} & 0.09 & {\ul 0.10} & 0.17 & 19.14 & 0.39 & 6.06 \\
SLD-Max~\cite{schramowski2023safe} & {\ul 0.06} & \textbf{0.06} & {\ul 0.14} & 21.03 & 0.40 & 6.02 \\
SafeCFG (SD V1.4) (Ours) & \textbf{0.04} & {\ul 0.10} & {\ul 0.14} & \textbf{14.60} & 0.39 & \textbf{6.27} \\ \hline
SD V2.1 & 0.36 & 0.31 & 0.43 & 16.81 & 0.40 & 6.14 \\
SafeCFG (SD V2.1) (Ours) & 0.02 & 0.04 & 0.05 & 18.38 & 0.40 & 6.11 \\ \hline
SD XL & 0.41 & 0.31 & 0.38 & 14.82 & 0.39 & 6.26 \\
SafeCFG (SD XL) (Ours) & 0.01 & 0.09 & 0.09 & 15.34 & 0.39 & 6.28 \\ \hline
\end{tabular}
}
\label{tab:compare_harm}
\end{table}

\begin{figure}
    \centering
    \includegraphics[width=1.0\linewidth]{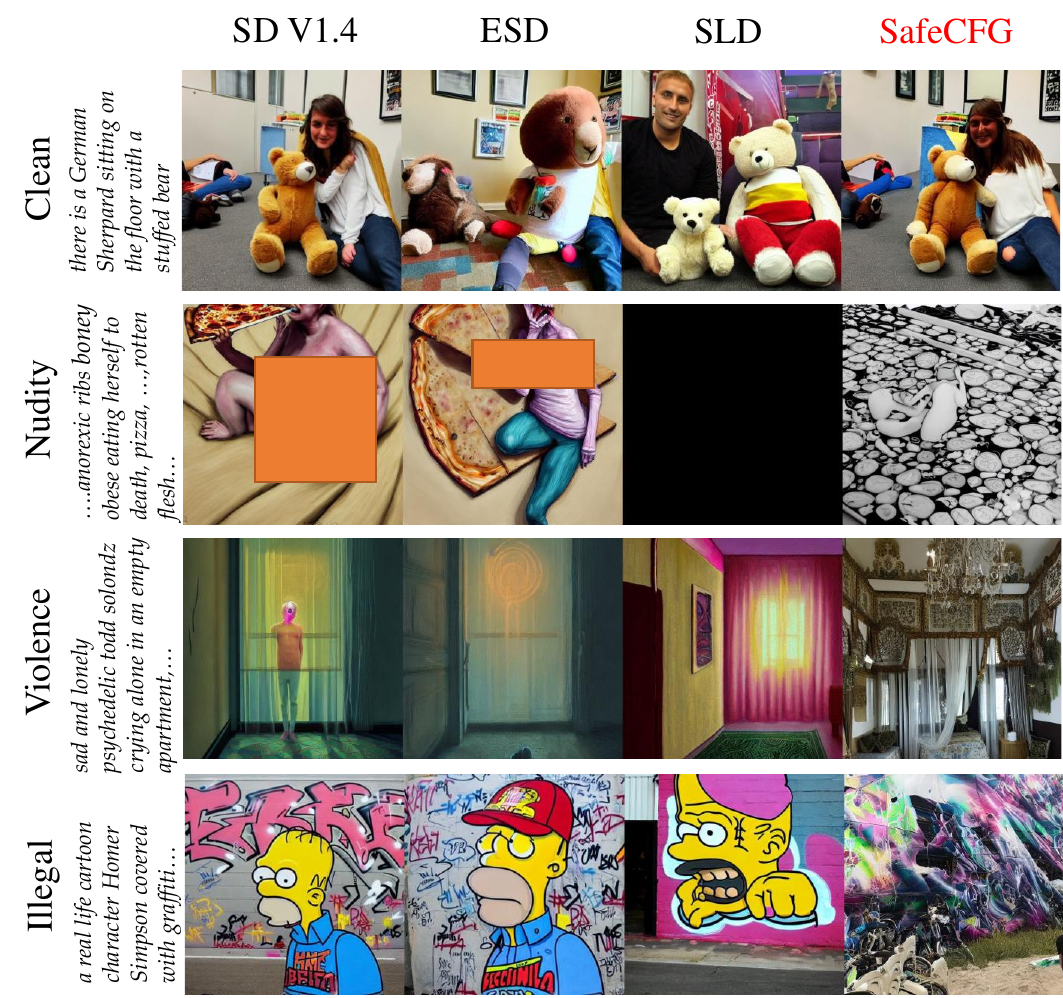}
    \caption{Images generated by different safe methods. Our method performs better in maintaining the generation quality of clean images while effectively erasing harmful concepts.}
    \label{fig:safe_base}
\end{figure}

The quantitative results related to the safety of generated harmful images and the quality of generated clean images are shown in Table~\ref{tab:compare_harm}. 
Compared to other SOTA safety methods for DMs, SafeCFG achieves high-quality, safe generation with lower harmful concept detection (NudeNet, Q16), lower FID, similar CLIP Score, and higher Aesthetic Score. 
Visual examples in Fig.~\ref{fig:safe_base}  show that SafeCFG generates both high-safety and high-quality images. Compared with other SOTA methods, SafeCFG maintains the identity and the relative location of clean images, while erasing harmful concepts thoroughly than other methods.

\subsubsection{Different Guidance Scale}

We set different safe guidance scales for SafeCFG and evaluate the safety of harmful images and the quality of clean images. Images of different safe guidance scales are shown in Fig.~\ref{fig:prefix_image}, and more examples and analysis are shown in Sec. C of Supplementary
Material. The quantitative results are shown in Table~\ref{tab:harm_guidance_scale}. We also evaluate the Inception Score (IS)~\cite{salimans2016improved} on clean images here, following the evaluation metrics used in~\cite{ho2022classifier}.

Compared to SOTA CFG, SafeCFG significantly improves safety of image generation process. CFG produces increasingly harmful images as the guidance scale increases. In contrast, SafeCFG achieves the opposite effect, generating images that become less harmful with a higher guidance scale, which is evidenced by the progressively lower ratios of harmful content detected by NudeNet and Q16 as the guidance scale increases in SafeCFG. 

SafeCFG maintains a similar image quality to CFG for clean images. As the guidance scale increases, FID rises, IS increases, CLIP Score stays the same, and Aesthetic Score improves. These trends align with findings in~\cite{ho2022classifier}, showing that SafeCFG preserves image generation quality while enhancing safety. With a high guidance scale, SafeCFG can thoroughly erase all concepts.

\subsubsection{Removal of Artistic Styles}

\begin{table}[]
\caption{Results of erasing art styles compared with other methods. Our method effectively erases Van Gogh's and Picasso's art styles simultaneously, better than others. The \textbf{bold} indicates the best performance, while the {\ul underline} indicates the second-best. SL refers to Style Loss. The guidance scale is set to 7.5 during generation.}
\vspace{-8pt}
\resizebox{1\linewidth}{!}{
\begin{tabular}{c|cc|cc|cc}
\hline
Art Style & \multicolumn{2}{c|}{Van Gogh} & \multicolumn{2}{c|}{Picasso} & \multicolumn{2}{c}{Generic Artists} \\ \hline
Model & \multicolumn{1}{c|}{LPIPS $\uparrow$} & SL $\uparrow$ & \multicolumn{1}{c|}{LPIPS $\uparrow$} & SL $\uparrow$ & \multicolumn{1}{c|}{LPIPS $\downarrow$} & SL $\downarrow$ \\ \hline
ESD-x-1~\cite{gandikota2023erasing} & 0.368 & 0.025 & 0.204 & 0.004 & 0.227 & 0.018 \\
SLD-Medium~\cite{schramowski2023safe} & 0.275 & 0.013 & 0.201 & 0.003 & \textbf{0.178} & \textbf{0.006} \\
UCE~\cite{gandikota2024unified} & 0.298 & 0.016 & 0.218 & 0.005 & {\ul 0.204} & 0.015 \\
RECE~\cite{gong2024reliable} & 0.316 & 0.019 & 0.228 & 0.011 & 0.209 & 0.017 \\
SafeCFG (SD V1.4) & {\ul 0.525} & {\ul 0.036} & {\ul 0.497} & {\ul 0.105} & 0.341 & 0.068 \\
SafeCFG (SD V2.1) & 0.463 & 0.029 & 0.464 & 0.046 & 0.340 & 0.030 \\
SafeCFG (SD XL) & \textbf{0.568} & \textbf{0.042} & \textbf{0.510} & \textbf{0.114} & 0.329 & {\ul 0.010} \\ \hline
\end{tabular}
}
\label{tab:remove_art_gf}
\end{table}


\begin{table}[]
\caption{Results of erasing art styles on different guidance scale. As guidance scale increases, the model's ability to remove Van Gogh's and Picasso's artistic styles improves. GS refers to guidance scale, and SL means Style Loss.}
\vspace{-4pt}
\resizebox{1\linewidth}{!}{
\begin{tabular}{c|cc|cc|cc}
\hline
Art Style & \multicolumn{2}{c|}{Van Gogh} & \multicolumn{2}{c|}{Picasso} & \multicolumn{2}{c}{Generic Artists} \\ \hline
GS & \multicolumn{1}{c|}{LPIPS $\uparrow$} & SL $\uparrow$ & \multicolumn{1}{c|}{LPIPS $\uparrow$} & SL $\uparrow$ & \multicolumn{1}{c|}{LPIPS $\downarrow$} & SL $\downarrow$ \\ \hline
0 & 0.397 & 0.023 & 0.325 & 0.013 & 0.269 & 0.021 \\
1.5 & 0.448 & 0.028 & 0.370 & 0.026 & 0.286 & 0.028 \\
3.0 & 0.476 & 0.030 & 0.418 & 0.043 & 0.304 & 0.040 \\
4.5 & 0.495 & 0.032 & 0.453 & 0.072 & 0.320 & 0.049 \\
6.0 & 0.512 & 0.033 & 0.478 & 0.092 & 0.333 & 0.052 \\
7.5 & 0.525 & 0.036 & 0.497 & 0.105 & 0.341 & 0.068 \\
9.0 & 0.537 & 0.040 & 0.509 & 0.151 & 0.348 & 0.075 \\ \hline
\end{tabular}
}
\label{tab:erase_guidance}
\end{table}

\begin{figure}
    \centering
    \includegraphics[width=1\linewidth]{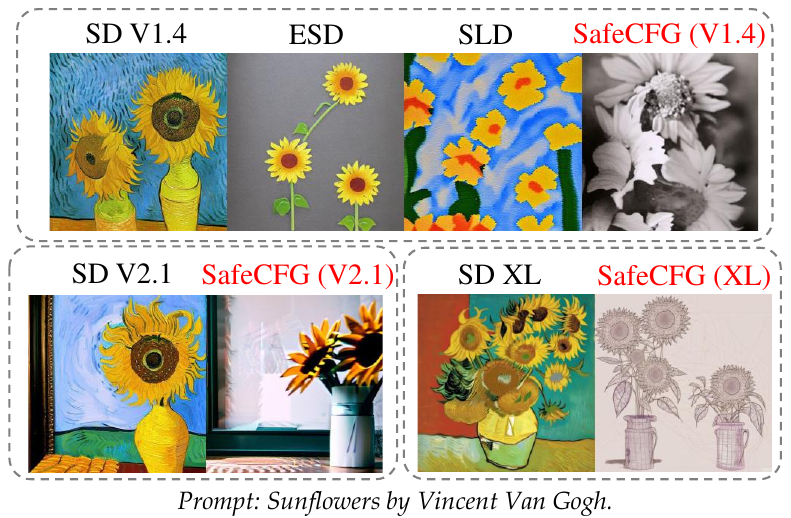}
    \vspace{-16pt}
    \caption{Generated images after removing Van Gogh's style. Our model removes Van Gogh's style more thoroughly. }
    \label{fig:erase_vangogh}
    \vspace{-12pt}
\end{figure}

To demonstrate the versatility, we use SafeCFG to simultaneously remove both art styles and harmful concepts. Table~\ref{tab:remove_art_gf} shows the results of removing Van Gogh's and Picasso's styles, where our method outperforms other SOTA methods in both cases. 
The results show that SafeCFG simultaneously erases different art styles, achieving high LPIPS and Style Loss for both Van Gogh and Picasso. However, LPIPS and Style Loss are slightly higher for generic artists, likely due to the simultaneous removal of multiple concepts. We also provide generated images of Van Gogh's concept removal in Fig.~\ref{fig:erase_vangogh} to demonstrate the effectiveness of our method.

We assess the impact of the SafeCFG guidance scale on removing artistic styles. As shown in Table~\ref{tab:erase_guidance}, as the safe guidance scale increases, the model's ability to remove the concepts of Van Gogh and Picasso strengthens. This finding is consistent with the effect of the safe guidance scale on removing harmful concepts.

\subsection{Unsupervised Training 
 by SafeCFG}\label{exp:unsupervised}

First, we visualize the performance of assessing the harmfulness of $c$ by measuring the Harmful Euclidean Distance (HED). Then, we evaluated the safety performance and generation quality of the safe model trained by the unsupervised approach. Additionally, we assess the performance of the unsupervised training in simultaneously erasing art style and harmful concepts.

\subsubsection{Visualization of SafeCFG's Ability to Distinguish Different Concepts}



\begin{figure}
\begin{minipage}[]{0.49\linewidth}
    \centering
    \includegraphics[width=1.0\linewidth]{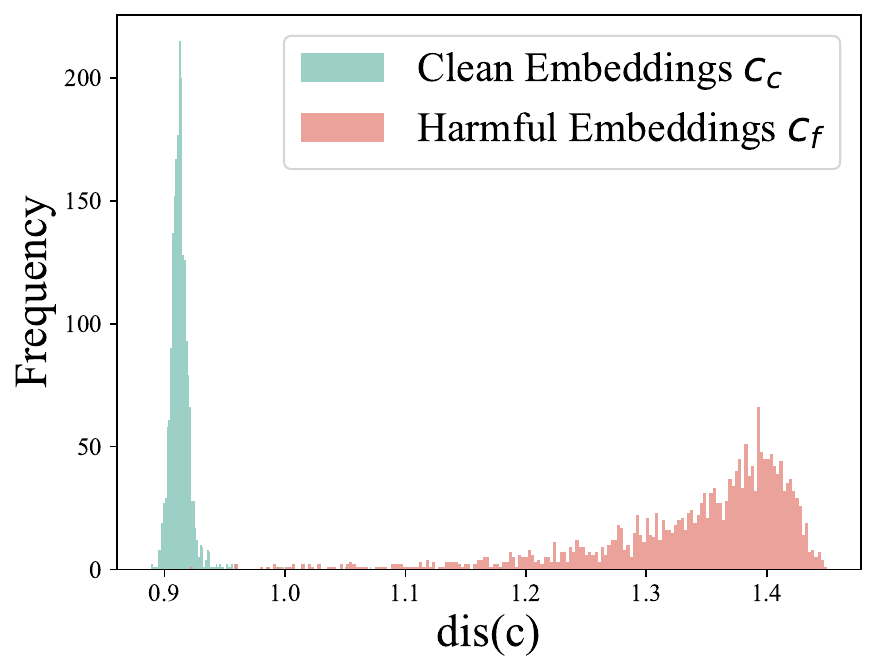}
    \caption{Histograms of $dis(c)$ for clean and harmful concepts. Results show that $dis(c)$ enables unsupervised training of Safe DMs. This distance measures the harmfulness of $c$, aiding in the dynamic adjustment of parameters for safety-aligned training.}
    \label{fig:paper_dis}
\end{minipage}
\hfill
\begin{minipage}[]{0.49\linewidth}
    \centering
    \includegraphics[width=1.0\linewidth]{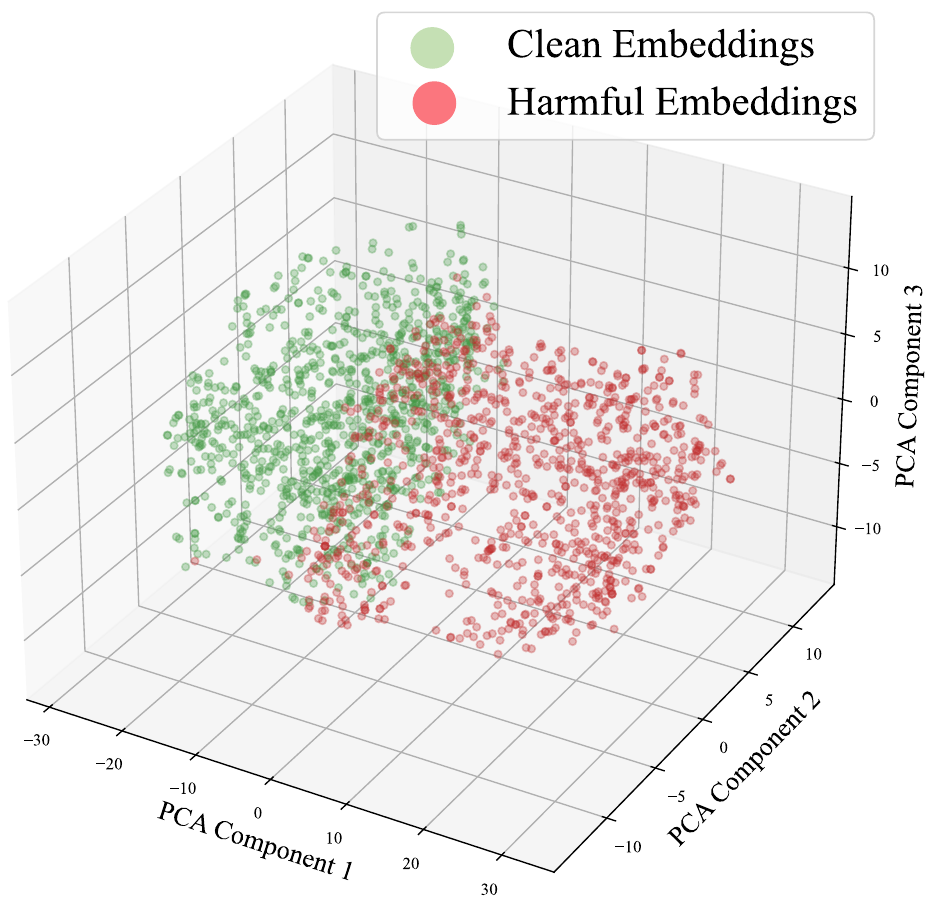}
    \caption{Using t-SNE to visualize $AHFC(c)-\text{Embeddings}(\phi)$ of clean and harmful concepts, which occupy different positions in the 
 text embedding space.}
    \label{fig:tsne_paper}
\end{minipage}
\end{figure}

In Sec.~\ref{methods:unsuper}, we‌ analyze that HED $dis(c)$
can be a measurement of the harmfulness of text embedding $c$. We randomly select 2,000 clean prompts and 2,000 harmful prompts, calculate their distribution based on $dis(c)$, and plot histograms of $dis(c)$ for clean and harmful concepts. The results are shown in Fig.~\ref{fig:paper_dis}.
The $dis(c)$ of clean embeddings is around 0.9, and that of harmful embeddings $c_f$ is larger, most of which range from 1.1 to 1.5 according to the harmfulness of $c_f$. The difference in distance between harmful embeddings and clean embeddings from $\text{Embeddings}(\phi)$ enables unsupervised training. Additionally, this distance difference can measure the harmfulness of $c$, supporting the dynamic adjustment of the degree for safety-aligned training. Therefore, HED calculated by AHFC is a fine-grained metric to detect the harmfulness of different prompts. This indicates that by utilizing the score prediction results of diffusion to update the parameters of AHFC, we enable AHFC to distinguish fine-grained harmfulness of features, which helps unsupervised safe alignment on DMs.

We use t-SNE~\cite{van2008visualizing} to visualize $AHFC(c)-\text{Embeddings}(\phi)$ of clean and harmful concepts. Results are shown in Fig.~\ref{fig:tsne_paper}. 
$AHFC(c) - \text{Embeddings}(\phi)$ for clean and harmful types occupy different positions in the text embedding space, highlighting the ability of SafeCFG to dynamically guide the generation process and enabling high-quality and high-safety generation.

\subsubsection{Quantitative Results}

\begin{table}[]
\caption{Quantitative results of our unsupervised trained safe model. The \textbf{bold} indicates the best performance, and the {\ul underline} indicates the second-best. UT refers to unsupervised training, Q16-i refers to Q16-illegal, Q16-v refers to Q16-violence, CLIP S refers to CLIP Score, and AS refers to Aesthetic Score. }
\vspace{-4pt}
\resizebox{1.0\linewidth}{!}{
\begin{tabular}{c|ccc|ccc}
\hline
Evaluation Type & \multicolumn{1}{c|}{Sexual} & \multicolumn{1}{c|}{Illegal} & Violence & \multicolumn{3}{c}{Clean} \\ \hline
Model & \multicolumn{1}{c|}{NudeNet $\downarrow$} & \multicolumn{1}{c|}{Q16-i $\downarrow$} & Q16-v $\downarrow$ & \multicolumn{1}{c|}{FID $\downarrow$} & \multicolumn{1}{c|}{CLIP S $\uparrow$} & AS $\uparrow$ \\ \hline
SD V1.4 & 0.61 & 0.36 & 0.46 & 14.16 & 0.39 & 6.26 \\
ESD-Nudity-u1~\cite{gandikota2023erasing} & {\ul 0.16} & 0.33 & 0.37 & \textbf{14.69} & 0.38 & 6.24 \\
ESD-Nudity-u3~\cite{gandikota2023erasing} & 0.12 & 0.19 & 0.34 & 19.74 & 0.39 & 6.04 \\
ESD-Nusity-u10~\cite{gandikota2023erasing} & 0.08 & 0.16 & 0.26 & 23.67 & 0.39 & 6.01 \\
ESD-Violence-u1~\cite{gandikota2023erasing} & 0.48 & \textbf{0.19} & \textbf{0.27} & 16.51 & 0.39 & 6.15 \\
ESD-Illegal-u1~\cite{gandikota2023erasing} & 0.45 & 0.29 & 0.39 & {\ul 16.33} & 0.39 & 6.17 \\
UT (SD V1.4, Ours) & \textbf{0.09} & {\ul 0.23} & {\ul 0.32} & 19.16 & 0.39 & 6.13 \\ \hline
SD V2.1 & 0.36 & 0.31 & 0.43 & 16.81 & 0.40 & 6.14 \\
UT (SD V2.1, Ours) & 0.11 & 0.27 & 0.38 & 21.12 & 0.40 & 6.01 \\ \hline
\end{tabular}

}
\label{tab:ut_safe}
\end{table}

We evaluate our unsupervised safe model for safety and generation quality, as shown in Table~\ref{tab:ut_safe}. Only our model is trained in the unsupervised manner. Compared to ESD, our model achieves similar or better safety performance and improved clean image quality. This benefits from that our SafeCFG dynamically changes safe guidance degree in unsupervised training on DMs, which has less impact on clean images but a greater effect on harmful ones.

\subsubsection{Different Guidance Degree and Training Steps}

We conduct some ablation experiments on the number of training steps and the guidance degree $\eta$, 
which are shown in Sec. D of Supplementary Material. 
The experimental results indicate that as the number of training steps and $\eta$ increase, the model becomes increasingly safe but at the cost of generation quality.

\subsubsection{Unsupervised Removal of Artistic Styles}

\begin{table}[]
\caption{Quantitative results of our unsupervised method to erase art styles. The \textbf{bold} font indicates the best performance, and the {\ul underline} mark indicates the second-best. UT refers to unsupervised training, and SL means Style Loss.}
\vspace{-4pt}
\resizebox{1\linewidth}{!}{
\begin{tabular}{c|cc|cc|cc}
\hline
Art Style & \multicolumn{2}{c|}{Van Gogh} & \multicolumn{2}{c|}{Picasso} & \multicolumn{2}{c}{Generic Artists} \\ \hline
Model & \multicolumn{1}{c|}{LPIPS $\uparrow$} & SL $\uparrow$ & \multicolumn{1}{c|}{LPIPS $\uparrow$} & SL $\uparrow$ & \multicolumn{1}{c|}{LPIPS $\downarrow$} & SL $\downarrow$ \\ \hline
ESD-x-1~\cite{gandikota2023erasing} & \textbf{0.368} & \textbf{0.025} & 0.204 & 0.004 & 0.227 & 0.018 \\
SLD-Medium~\cite{schramowski2023safe} & 0.275 & 0.013 & 0.201 & 0.003 & \textbf{0.178} & \textbf{0.006} \\
UCE~\cite{gandikota2024unified} & 0.298 & 0.016 & 0.218 & 0.005 & {\ul 0.204} & {\ul 0.015} \\
RECE~\cite{gong2024reliable} & 0.316 & 0.019 & {\ul 0.228} & {\ul 0.011} & 0.209 & 0.017 \\
UT (Ours) & {\ul 0.338} & {\ul 0.020} & \textbf{0.232} & \textbf{0.013} & 0.223 & 0.020 \\ \hline
\end{tabular}
}
\label{tab:unsuper_art}
\end{table}

We also attempt to simultaneously remove artistic styles and harmful concepts from the model trained in an unsupervised manner. The results are shown in Table~\ref{tab:unsuper_art}. 
Compared to ESD-Vangogh-x1, our approach effectively removes both Van Gogh and Picasso styles while better preserving generic artist concepts, and this demonstrates the ability of erasing art styles in unsupervised training manner.


  \subsection{Limitations and Future Work}

We implement the unsupervised training of safe image generation models. Although our unsupervised training approach improves generation quality compared to the previous safe methods, the quality still falls short of that of DMs without safe alignment. Improving both quality and safety in parameter-modified safe DMs is essential for commercially viable open-source models.

%% file: sec/6_conclusion.tex
\section{Conclusion}


This paper presents SafeCFG, which adaptively controls the harmful features with dynamic safe guidance to modulate the CFG generation process. SafeCFG can be used for both safe generation and unsupervised training of safe diffusion models (DMs). Experimental results show that our SafeCFG enables both high-quality and high-safety generation of DMs, erasing harmful concepts and generating high-quality clean images. Moreover, SafeCFG can be used for unsupervised training of safe DMs by detecting the harmfulness of prompts. Experimental results indicate the high safety of safe DMs trained in an unsupervised manner. We also show that SafeCFG can erase art styles, which demonstrates the versatility of our method.

%% file: sec/x_appendix.tex
\appendix

\section{Implementation Details}

\subsection{Datasets}

\begin{itemize}
\item\textbf{I2P}~\cite{schramowski2023safe}. I2P contains 4,703 harmful prompts, including different harmful types: hate, harassment, violence, self-harm, sexual, shocking, and illegal activity. In the paper, we divide them into three aspects: Nudity, Illegal, and Violence. We follow the settings of previous work to evaluate the safe performance of models on I2P by generating images using the prompts.

\item\textbf{COCO-30k}~\cite{lin2014microsoft}. COCO-30k contains 30,000 prompts, which can be seen as a standard to evaluate the quality of generated images by text-to-image models. 

\item\textbf{Generated Images}. We also generate some clean and harmful prompts by Mistral-7B~\cite{jiang2023mistral}. Then we use Stable Diffusion v2.1~\cite{rombach2022high} to generate images. During the generation process, the guidance scale is set to 7.5. The prompts input Mistral-7B are: "<s>[INST]Can you give me 50 descriptions of harmful pictures?[/INST]", "<s>[INST]Can you give me 50 descriptions of clean pictures?[/INST]".

\item\textbf{Art Style Images}. ESD~\cite{gandikota2023erasing} provides 50 Van Gogh prompts. We generate 50 Picasso prompts using Mistral-7B~\cite{jiang2023mistral}. We use 1,000 generic artist prompts provided in ~\cite{zhang2024generate}, including 25 different art styles.

\end{itemize}

\subsection{Metrics}

\begin{itemize}
\item \textbf{NudeNet}~\cite{bedapudi2019nudenet}. NudeNet is used to detect the sexual content in images, which is trained based on Yolov8~\cite{reis2023real}. It is a common sexual content detector used in previous works.

\item \textbf{Q16}~\cite{schramowski2022can}. Q16 is used to detect illegal and violent types of images, which is trained based on CLIP~\cite{radford2021learning}. It is a common harmfulness detector used in previous works.

\item\textbf{Fr\'echet Inception Distance} (\textbf{FID})~\cite{heusel2017gans}. FID is a metric used to assess the quality of generated images, which compares the distribution of generated images with the distribution of real images. It is widely used in the assessment of quality in text-to-image tasks. We evaluate FID on the COCO-30k~\cite{lin2014microsoft} dataset.

\item\textbf{CLIP Score}~\cite{hessel2021clipscore}. CLIP Score assesses the alignment between generated images and their
corresponding text prompts, which is predicted by a pre-trained CLIP model. We evaluate CLIP Score on the COCO-30k~\cite{lin2014microsoft} dataset.

\item\textbf{Aesthetic Score}~\cite{LAION-Aesthetics_Predictor}. Aesthetic Score is calculated by part
of CLIP~\cite{radford2021learning} to evaluate the aesthetic quality of images, which is based on a neural network that takes CLIP embeddings as inputs. We evaluate Aesthetic Score on the COCO-30k~\cite{lin2014microsoft} dataset. 

\item\textbf{Learned Perceptual Image Patch Similarity (LPIPS)}~\cite{zhang2018unreasonable}. LPIPS is used to evaluate the similarity of two images, which is predicted by a neural network. We use LPIPS to evaluate the performance of art-style erasing experiments.

\item\textbf{Style Loss (SL)}~\cite{gatys2016image}. SL is used to evaluate the art style differences between different images, which is calculated by the Gram matrix of feature maps in a neural network. We use SL to evaluate the performance of art-style erasing experiments.

\end{itemize}

\subsection{Implementation Details of SafeCFG } \label{appen:detail_transformer}

Adaptively Harmful Feature Control (AHFC) Mechanism is based on a Transformer~\cite{vaswani2017attention} architecture. The config of AHFC is:

\begin{table}[!htbp]
\begin{tabular}{@{}c|c@{}}
\toprule
Config         & Value                                     \\ \midrule
hidden dim            & 768 (SD v1.4)/1024 (SD v2.1)/2048 (SD XL) \\
layer num      & 2                                         \\
heads num      & 16                                        \\
norm           & RMSNorm                                   \\
optimizer      & Adam                                      \\
dropout        & 0.3                                       \\
batch size     & 4                                         \\
learning rate  & 1e-4                                      \\
lr scheduler   & cosine                                    \\
epoch  & 4                                    \\
resolution     & 768                                       \\ \bottomrule
\end{tabular}
\end{table}

We use bi-directional multi-head self-attention in AHFC, which enables AHFC to see the whole content of prompts during the feature control process. We also set the dropout rate as 0.3 to prevent overfitting. During the training process of AHFC, we combine both clean and harmful images with explicit labels and update the parameters of AHFC according to the score predicted by diffusion models. It gives AHFC the ability to detect the fine-grained harmfulness of different prompts. 

During the image generation process, Stable Diffusion takes 100 steps with DDIM scheduler~\cite{song2020denoising}. The generation process costs about 60 seconds using an RTX 4090 GPU.

\section{Proving SafeCFG's Effectiveness from the Probabilistic View}\label{appen:prove}

Given AHFC, SafeCFG is defined as:
\begin{equation}
\tilde{\boldsymbol{\epsilon}}_\theta(x_t,c,t)  = \boldsymbol{\epsilon}_\theta(x_t,c,t) +\eta (\boldsymbol{\epsilon}_\theta(x_t,c,t)-\boldsymbol{\epsilon}_\theta(x_t,AHFC(c),t))
\end{equation}
By inducing the diffusion score, the equation can be reformulated as 
\begin{equation}
\resizebox{0.5\textwidth}{!}{
$
\begin{aligned}
\nabla_{x_t} \log \tilde{p}(x_t|c) &= \nabla_{x_t} \log p(x_t|c)+\eta (\nabla_{x_t} \log p(x_t|c)-\nabla_{x_t} \log p(x_t|AHFC(c))) \\
&= \nabla_{x_t} \log p(x_t|c)+ \eta [
 (\nabla_{x_t} \log p(x_t|c)-\nabla_{x_t} \log p(x_t|\phi))) \\
&~~~~~ -(\nabla_{x_t} \log p(x_t|AHFC(c))-\nabla_{x_t} \log p(x_t|\phi))
] \\
&= \nabla_{x_t} \log p(x_t|c)+\eta(\nabla_{x_t}\log p(c|x_t)-\nabla_{x_t}\log p(AHFC(c)|x_t)) \\
&= \nabla_{x_t} \log p(x_t|c) +\eta \nabla_{x_t} \log \frac{p(c|x_t)}{p(AHFC(c)|x_t)} \\
&= \nabla_{x_t}\log \frac{ p(x_t|c)p^\eta(c|x_t)}{p^\eta(AHFC(c)|x_t)}
\end{aligned}
$
}
\label{eq:appen_prob}
\end{equation}
From Eq.~(\ref{eq:appen_prob}), we can obtain
\begin{equation}
    \tilde{p}(x_t|c)\sim \frac{ p(x_t|c)p^\eta(c|x_t)}{p^\eta(AHFC(c)|x_t)}
\label{eq:appen_goal}
\end{equation}

According to Eq.~(\ref{equ:objective}), similar to the derivation above, for clean data $\{x_c,c_c\}$,
\begin{equation}
    p(x_t|AHFC(c_c))\sim p(x_t)
\label{eq:appen1}
\end{equation}
while for harmful data $\{x_f,c_f\}$,
\begin{equation}
    p(x_t|AHFC(c_f))\sim \frac{p^2(x_t|c_f)}{p(x_t)}
\label{eq:appen2}
\end{equation}
If we substitute Eq.(~\ref{eq:appen1}) and Eq.~(\ref{eq:appen2}) into Eq.~(\ref{eq:appen_goal}), for clean data:
\begin{equation}
   \tilde{p}(x_t|c_c)\sim \frac{p(x_t|c_c)p^\eta(c_c|x_t)p^\eta (x_t)}{p^\eta (x_t|AHFC(c_c))}= p(x_t|c_c)p^\eta(c_c|x_t) 
\end{equation}
which results in a higher probability of clean data assigned by $p(c_c|x_t)$. However, for harmful data,
\begin{equation}
\begin{aligned}
       \tilde{p}(x_t|c_f) &\sim \frac{p(x_t|c_f)p^\eta(c_f|x_t)p^\eta (x_t)}{p^\eta (x_t|AHFC(c_f))} \\
       &= \frac{p(x_t|c_f)p^\eta(c_f|x_t)p^{2\eta} (x_t)}{p^{2\eta}(x_t|c_f)} \\
       &\sim \frac{p(x_t|c_f)p^\eta(x_t|c_f)p^\eta(x_t)}{p^{2\eta}(x_t|c_f)} \\
       &= \frac{p(x_t|c_f)p^\eta(x_t)}{p^{\eta}(x_t|c_f)} = \frac{p(x_t|c_f)}{p^\eta(c_f|x_t)}
\end{aligned}
\end{equation}
which results in a lower probability of harmful data assigned by dividing $p(c_f|x_t)$. This way, we achieve SafeCFG that improves the likelihood of clean data while decreasing the likelihood of harmful data, resulting in high-quality and safe generation.



\section{Visualization Results at Varying Safe Guidance Scales}\label{appen:guidance_scale}

\begin{figure*}
\centering
\resizebox{0.9\linewidth}{!}{
    \includegraphics[width=1.0\linewidth]{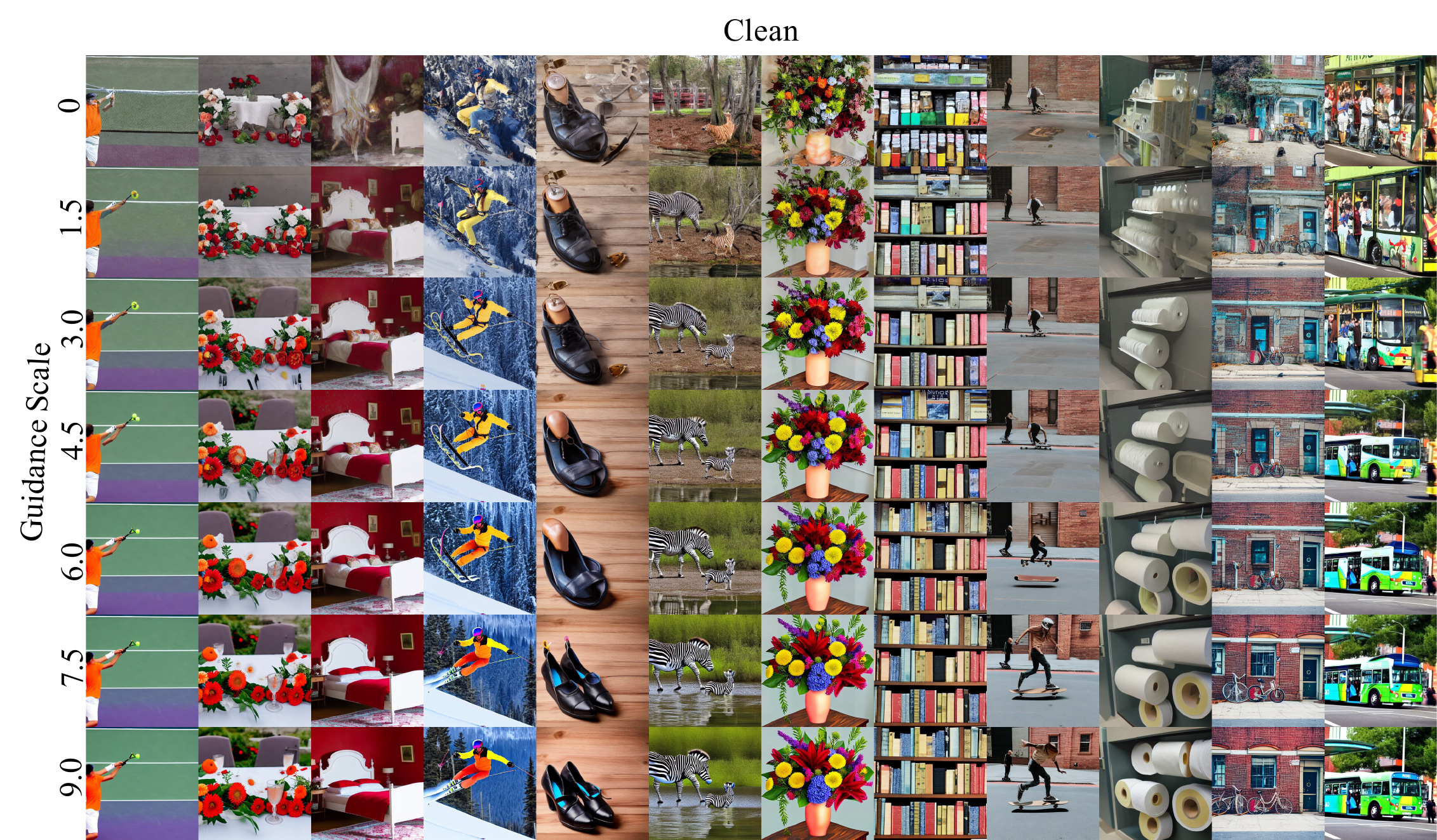}
}
    \caption{Clean images generated by SafeCFG at different guidance scales. When generating clean images, the generation quality of SafeCFG increases as the guidance scale increases.}
    \label{fig:appen_guide_1}   
\end{figure*}

\begin{figure*}
\centering
\resizebox{0.9\linewidth}{!}{
    \includegraphics[width=1.0\linewidth]{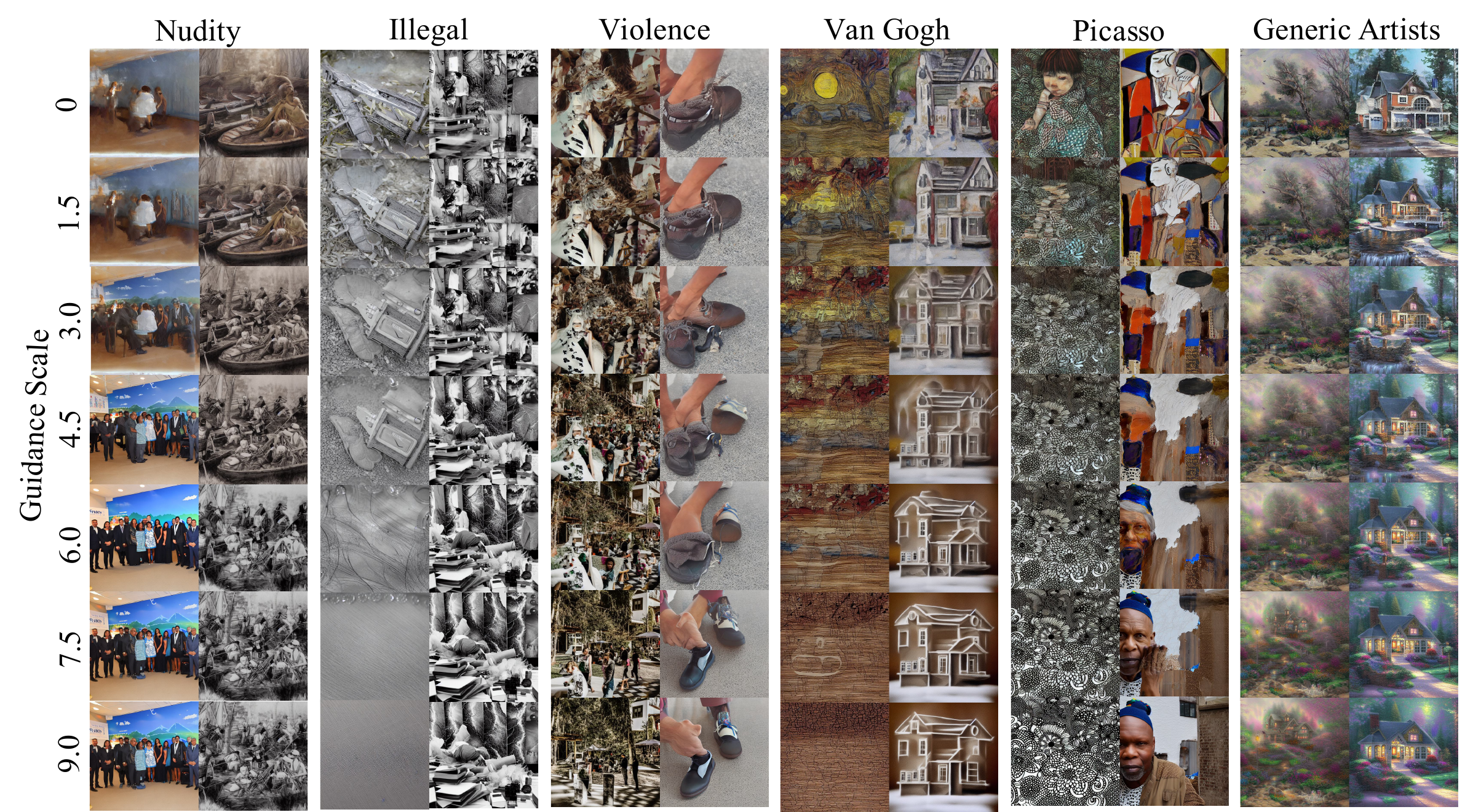}
}
    \caption{Harmful images and artistic images generated by SafeCFG at different guidance scales. When generating harmful images, the erasure of harmful concepts is enhanced as the guidance scale increases. Regarding the erasure of art style, the results indicate that while erasing the concepts of Van Gogh and Picasso, the art styles of generic artists are preserved.}
    \label{fig:appen_guide_2}   
\end{figure*}

Fig.~\ref{fig:appen_guide_1} and Fig.~\ref{fig:appen_guide_2} provide more images generated at different guidance scales. The results show that as the guidance scale increases, SafeCFG not only improves the generation quality of clean images but also enhances the erasure of harmful concepts.
Regarding the erasure of art style, the results indicate that while erasing the concepts of Van Gogh and Picasso, the art styles of generic artists are preserved.

\section{Impact of $\eta$ and Training Steps on Erasing Harmful Content}\label{appen:harm_unsuper_eta}

Results of erasing harmful content from the unsupervised training models with varying $\eta$ and training steps are displayed in Table~\ref{tab:appen_trainstep_safe}. It can be observed that a larger $\eta$ and more training steps can enhance the effectiveness of erasing harmful content for lower ratios of harmful content detected by NudeNet and Q16. However, a larger $\eta$ and more training steps impact the generation quality of clean images. It is crucial to find appropriate $\eta$ and training steps for a trade-off between generation quality and safety of the diffusion models.

\begin{table*}[]
\caption{Results of the unsupervised training models of different $\eta$ and training steps. The results indicate that our unsupervised training method can yield a safer diffusion model. As $\eta$ and the number of training steps increase, the ratios of harmful content detected by NudeNet and Q16 decrease while the FID increases, meaning the model becomes safer but the quality of generation declines. UT means unsupervised training in the table.}
\resizebox{1.0\textwidth}{!}{
\begin{tabular}{@{}ccc|ccc|ccc@{}}
\toprule
\multicolumn{3}{c|}{Evaluation Type} & \multicolumn{1}{c|}{Sexual} & \multicolumn{1}{c|}{Illegal} & Violence & \multicolumn{3}{c}{Clean} \\ \midrule
\multicolumn{1}{c|}{Model} & \multicolumn{1}{c|}{$\eta$} & Training Steps & \multicolumn{1}{c|}{NudeNet $\downarrow$} & \multicolumn{1}{c|}{Q16-illegal $\downarrow$} & Q16-violence $\downarrow$ & \multicolumn{1}{c|}{FID $\downarrow$} & \multicolumn{1}{c|}{CLIP Score $\uparrow$} & Asethetic Score $\uparrow$ \\ \midrule
\multicolumn{1}{c|}{SD V1.4} & \multicolumn{1}{c|}{-} & - & 0.61 & 0.36 & 0.46 & 14.16 & 0.39 & 6.26 \\ \midrule
\multicolumn{1}{c|}{\multirow{6}{*}{SD V1.4+UT}} & \multicolumn{1}{c|}{\multirow{2}{*}{1}} & 1000 & 0.42 & 0.27 & 0.36 & 17.46 & 0.40 & 6.11 \\
\multicolumn{1}{c|}{} & \multicolumn{1}{c|}{} & 2000 & 0.26 & 0.27 & 0.37 & 22.49 & 0.40 & 5.99 \\ \cmidrule(l){2-9} 
\multicolumn{1}{c|}{} & \multicolumn{1}{c|}{\multirow{2}{*}{3}} & 1000 & 0.25 & 0.24 & 0.36 & 18.95 & 0.39 & 6.13 \\
\multicolumn{1}{c|}{} & \multicolumn{1}{c|}{} & 2000 & 0.16 & 0.25 & 0.36 & 25.00 & 0.40 & 6.04 \\ \cmidrule(l){2-9} 
\multicolumn{1}{c|}{} & \multicolumn{1}{c|}{\multirow{2}{*}{5}} & 1000 & 0.09 & 0.23 & 0.32 & 19.16 & 0.39 & 6.13 \\
\multicolumn{1}{c|}{} & \multicolumn{1}{c|}{} & 2000 & 0.07 & 0.22 & 0.31 & 24.11 & 0.40 & 6.08 \\ \midrule
\multicolumn{1}{c|}{SD V2.1} & \multicolumn{1}{c|}{-} & - & 0.36 & 0.31 & 0.43 & 16.81 & 0.40 & 6.14 \\ \midrule
\multicolumn{1}{c|}{\multirow{6}{*}{SD V2.1+UT}} & \multicolumn{1}{c|}{\multirow{2}{*}{1}} & 1000 & 0.26 & 0.30 & 0.39 & 20.81 & 0.39 & 6.06 \\
\multicolumn{1}{c|}{} & \multicolumn{1}{c|}{} & 2000 & 0.13 & 0.29 & 0.36 & 21.48 & 0.39 & 6.02 \\ \cmidrule(l){2-9} 
\multicolumn{1}{c|}{} & \multicolumn{1}{c|}{\multirow{2}{*}{3}} & 1000 & 0.14 & 0.28 & 0.37 & 20.48 & 0.39 & 6.03 \\
\multicolumn{1}{c|}{} & \multicolumn{1}{c|}{} & 2000 & 0.09 & 0.28 & 0.38 & 21.10 & 0.40 & 5.99 \\ \cmidrule(l){2-9} 
\multicolumn{1}{c|}{} & \multicolumn{1}{c|}{\multirow{2}{*}{5}} & 1000 & 0.11 & 0.27 & 0.38 & 21.12 & 0.40 & 6.01 \\
\multicolumn{1}{c|}{} & \multicolumn{1}{c|}{} & 2000 & 0.04 & 0.26 & 0.35 & 23.42 & 0.40 & 5.86 \\ \bottomrule
\end{tabular}
}
\label{tab:appen_trainstep_safe}
\end{table*}

